\documentclass[10pt,twocolumn,letterpaper]{article}

\usepackage{cvpr}
\usepackage{times}
\usepackage{epsfig}
\usepackage{graphicx}
\usepackage{amsmath}
\usepackage{amssymb}
\usepackage{color}
\usepackage{colortbl}
\usepackage{bm}
\definecolor{grayDark}{gray}{0.95}
\definecolor{grayLight}{gray}{0.98}

\usepackage[pagebackref=true,breaklinks=true,letterpaper=true,colorlinks,bookmarks=false]{hyperref}

\cvprfinalcopy 

\def\cvprPaperID{3692} 

\ifcvprfinal\fi
\begin{document}

\title{RepNet: Weakly Supervised Training of an Adversarial Reprojection Network for 3D Human Pose Estimation}

\author{Bastian Wandt and Bodo Rosenhahn\\
Leibniz Universit\"at Hannover\\
Hannover, Germany\\
{\tt\small wandt@tnt.uni-hannover.de}
}

\maketitle

\begin{abstract}
This paper addresses the problem of 3D human pose estimation from single images.
While for a long time human skeletons were parameterized and fitted to the observation by satisfying a reprojection error, nowadays researchers directly use neural networks to infer the 3D pose from the observations. 
However, most of these approaches ignore the fact that a reprojection constraint has to be satisfied and are sensitive to overfitting.
We tackle the overfitting problem by ignoring 2D to 3D correspondences. 
This efficiently avoids a simple memorization of the training data and allows for a weakly supervised training.
One part of the proposed reprojection network (RepNet) learns a mapping from a distribution of 2D poses to a distribution of 3D poses using an adversarial training approach. 
Another part of the network estimates the camera.
This allows for the definition of a network layer that performs the reprojection of the estimated 3D pose back to 2D which results in a reprojection loss function.

Our experiments show that RepNet generalizes well to unknown data and outperforms state-of-the-art methods when applied to unseen data. Moreover, our implementation runs in real-time on a standard desktop PC.
\end{abstract}

\section{Introduction}
\begin{figure}
	\centering
	\includegraphics[width=0.45\textwidth]{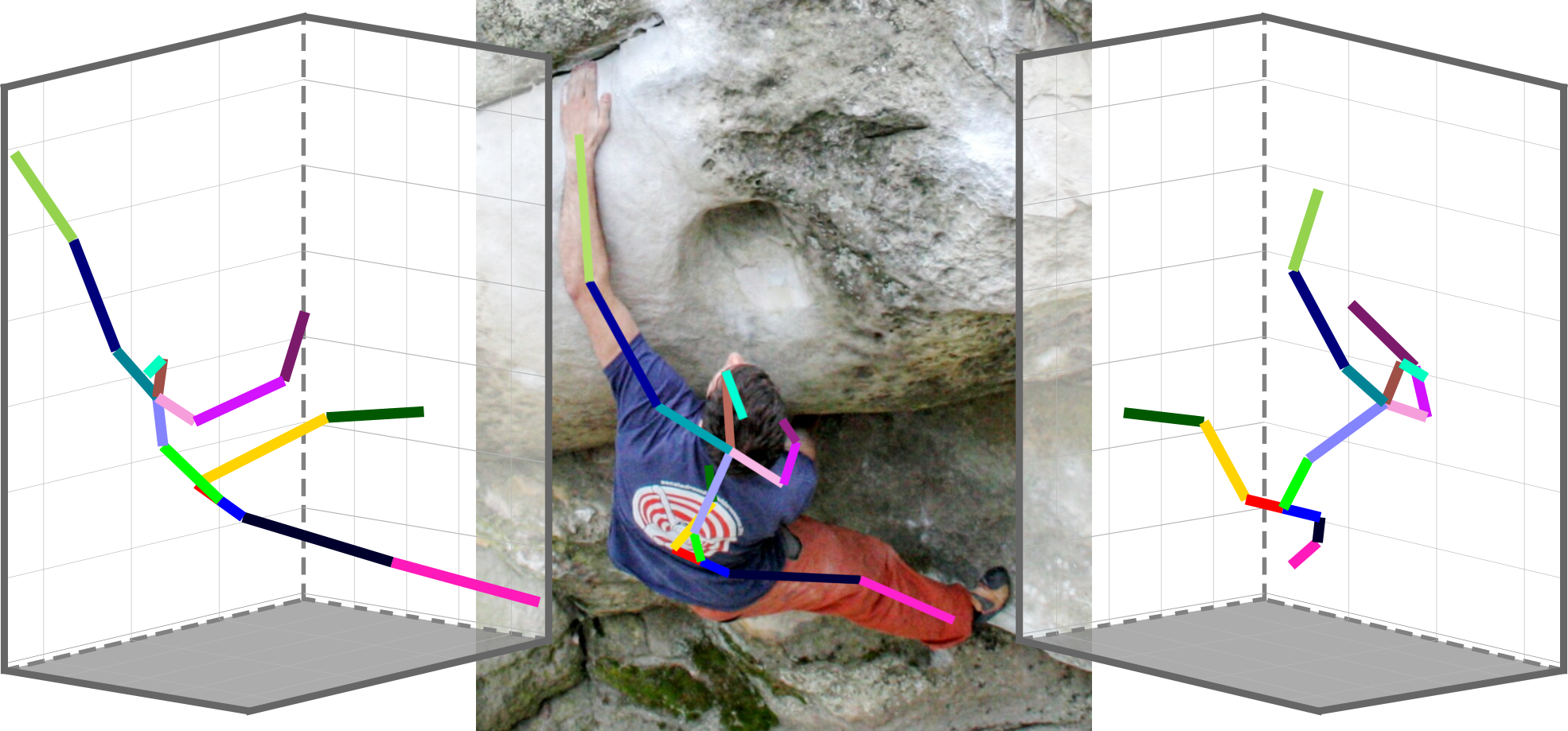}
	\caption{Our network predicts 3D human poses from noisy 2D joint detections. We use weakly supervised adversarial training without 2D to 3D point correspondences. Our critic networks enforces a plausible 3D pose while a reprojection layer projects the 3D pose back to 2D. Even strong deformations and unusal camera poses can be reconstructed.}
	\label{fig:teaser}
\end{figure}
\begin{figure*}[ht]
	\centering
	\includegraphics[width=0.85\textwidth]{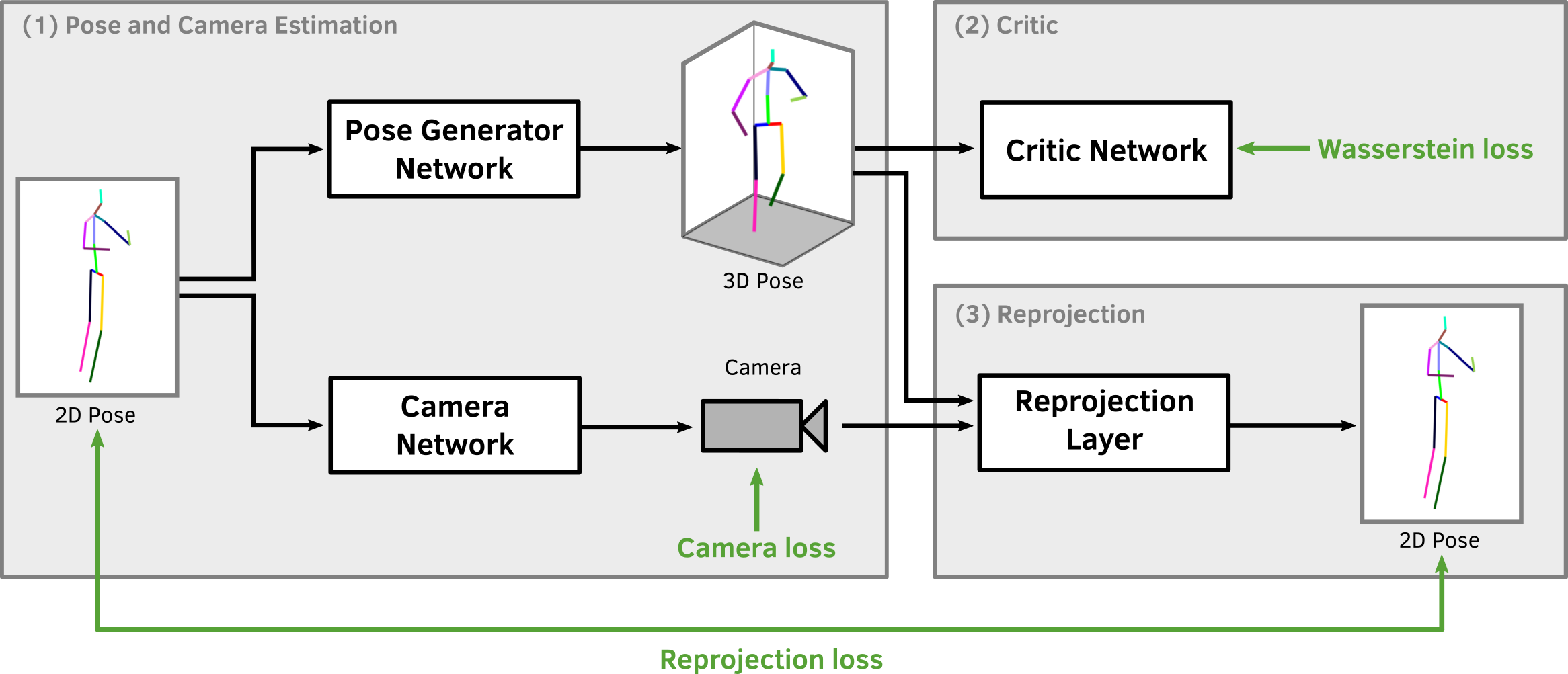}
	\caption{The proposed adversarial training structure for RepNet consist of three parts: a pose and camera estimation network (1), a critic network (2), and a reprojection network (3). There are losses (green) for the critic, the camera, and the reprojection.}
	\label{fig:network}
\end{figure*}

Human pose estimation from monocular images is a very active research field in computer vision with many  applications \eg in movies, medicine, surveillance, or human-computer interaction.
Recent approaches are able to infer 3D human poses from monocular images in good quality \cite{Park2016,Du2016,mpii3dhp2017,VNect_SIGGRAPH2017,Pavlakos2017,lcrnet2017,Moreno_cvpr2017,martinez_2017_3dbaseline,OriNet2018,Hossain2018}. 
However, most recent methods use neural networks that are straightforwardly trained with a strict assignment from input to output data \eg \cite{Park2016,Du2016,mpii3dhp2017,VNect_SIGGRAPH2017,Pavlakos2017,lcrnet2017,Moreno_cvpr2017,OriNet2018}.
This leads to surprisingly impressive results on similar data but usually the generalization to unknown motions and camera positions is problematic.
This paper presents a method to overcome this problem by using a neural network trained with a weakly supervised adversarial learning approach.
We relax the assumption that a specific 3D pose is given for every image in the training data by training a discriminator network --widely used in generative adversarial networks (GAN) \cite{Goodfellow2014}-- to learn a distribution of 3D human poses.
A second neural network learns a mapping from a distribution of detected 2D keypoints (obtained by \cite{StackedHourglassNewell2016}) to a distribution of 3D keypoints which are valid 3D human poses according to the discriminator network.
From the generative adversarial network point of view this can be seen as the generator network.
To force the generator network to generate matching 3D poses to the 2D observations we propose to add a third neural network that predicts camera parameters from the input data.
The inferred camera parameters are used to reproject the estimated 3D pose back to 2D which gives this framework its name: \textbf{Rep}rojection \textbf{Net}work (\textit{RepNet}).
Fig.~\ref{fig:network} shows an overview of the proposed network.
Additionally, to further enforce kinematic constraints we propose to employ an easy to calculate and implement descriptor for joint lengths and angles inspired by the kinematic chain space (KCS) of Wandt et al. \cite{WanAck2018a}.

In contrast to other works our proposed method is very robust against overfitting to a specific dataset.
This claim is reinforced by our experiments where the network can even infer human poses and camera positions that are not in the training set.
Even if there are strong deformations or unusual camera poses our network achieves good results as can be seen in the rock climbing image in Fig.~\ref{fig:teaser}.
This leads to our conclusion that the discriminator network does not \textit{memorize} all poses from the training set but learns a meaningful manifold of feasible human poses.
As we will show the inclusion of the KCS as a layer in the discriminator network plays an important role for the quality of the discriminator.

We evaluate our method on the three datasets Human3.6M \cite{h36m_pami}, MPI-INF-3DHP \cite{mpii3dhp2017}, and Leeds Sports Pose (LSP) \cite{LeadsSports2010}. 
On all the datasets our method achieves state-of-the-art results and even outperforms most supervised approaches.
Furthermore, the proposed network can predict a human pose in less than $0.1$ milliseconds on standard hardware which allows to build a real-time pose estimation system when combining it with state-of-the-art 2D joint detectors, such as OpenPose \cite{cao2017realtime}.

The code will be made available.

Summarizing, our contributions are:

\begin{itemize}
	\item An adversarial training method for a 3D human pose estimation neural network (RepNet) based on a 2D reprojection.
	\item Weakly supervised training without 2D-3D correspondences and unknown cameras.
	\item Simultaneous 3D skeletal keypoints and camera pose estimation.
	\item A layer encoding a kinematic chain representation that includes bone lengths and joint angle informations.
	\item A pose regression network that generalizes well to unknown human poses and cameras.
\end{itemize}

\section{Related Work}
The most relevant approaches related to our work can be roughly divided into two categories.
The first group consists of optimization-based approaches where a 3D human body model is deformed such that it satisfies a reprojection error.
The second group contains the most recent approaches that try to estimate 3D poses directly from images or detected keypoints.

\subsection{Reprojection Error Optimization}
Early works on human pose estimation from single images date back to Lee and Chen \cite{Lee1985} in 1985. 
They use known bone lengths and a binary decision tree to reconstruct a human pose.
Some authors \cite{Jiang2010,Gupta2014,ChenR17} propose to search for 3D poses in large pose databases that explain the 2D observations the best.
To compress the knowledge from these databases a widely used method is to learn an overcomplete dictionary of 3D human poses either using principal component analysis (PCA) or another dictionary learning method.
Commonly the best linear combination of bases obtained by a principal component analysis is optimized \cite{ChenC09,Wei2009,ZhouConvexRelax2016,Zhou2016}.
To constrain the optimization several priors are proposed, such as joint angle limits \cite{Akhter2015}, physical plausibility \cite{ZelWan2017}, or anthropometric regularization \cite{Ramakrishna12,SimoSerraRATM12,Wang2014}.
Other works enforce temporal coherence in video sequences \cite{Wandt2016, AllKas2017, WanAck2018a,ZelWan2017} or use additional sensors \cite{Marcard2016,Marcard2017,Marcard2018}.

\subsection{Direct Inference using Neural Networks}
Recently, many researchers focus on directly regressing the 3D pose from image data or from 2D detections using deep neural networks.
Several works try to build an end-to-end system which extracts the 3D pose from the image data \cite{Park2016,Du2016,mpii3dhp2017,VNect_SIGGRAPH2017,Pavlakos2017,lcrnet2017,OriNet2018,Kanazawa2018,NBF2018,pavlakos2018,Tung2017,Yang2018}.
Moreno-Noguer \cite{Moreno_cvpr2017} learns a mapping from 2D to 3D distance matrices.
Martinez et al. \cite{martinez_2017_3dbaseline} train a deep neural network on 2D joint detections to directly infer the 3D human pose.
They trained their network to achieve an impressive performance on the benchmark dataset Human3.6M \cite{h36m_pami}.
However, the network has significantly more parameters than poses in the training set of Human3.6M which could indicate a simple memorization of the training set.
Although our proposed pose estimation network has a similar number of parameters our experiments indicate that overfitting is avoided by our adversarial training approach.
Hossain et al. \cite{Hossain2018} extended the approach of \cite{martinez_2017_3dbaseline} by using a recurrent neural network for sequences of human poses.
The special case of weak supervision is rarely considered, 
Kanazawa et al. \cite{Kanazawa2018} propose a method that can also be trained without 2D to 3D supervision.
In contrast to our approach they use the complete image information to train an end-to-end model to reconstruct a volumetric mesh of a human body.
Yang et al. \cite{Yang2018} train a multi-source discriminator network to build an end-to-end model.

\section{Method}
The basic idea behind the proposed method is that 3D poses are regressed from 2D observations by learning a mapping from the input distribution (2D poses) to the output distribution (3D poses). 

In standard generative adversarial network (GAN) training \cite{Goodfellow2014} a generator network learns a mapping from an input distribution to the an output distribution which is rated by another neural network, called discriminator network.
The discriminator is trained to distinguish between real samples from a database and samples created from the generator network.
When training the generator to create samples that the discriminator predicts as real samples the discriminator parameters are fixed.
The generator and the discriminator are trained alternatingly and therefore compete with each other until they both converge to a minimum.

In standard GAN training the input is sampled from a gaussian or uniform distribution.
Here, we assume that the input is sampled from a distribution of 2D observations of human poses.
Adopting the Wasserstein GAN naming \cite{wgan2017} we call the discriminator \textit{critic} in the following. 
Without knowledge about camera projections the network produces random, yet feasible human 3D poses. 
However, these 3D poses are very likely the incorrect 3D reconstructions of the input 2D observations.
To obtain matching 2D and 3D poses we propose a camera estimation network followed by a reprojection layer.
As shown in Fig.~\ref{fig:network} the proposed network consists of three parts: The pose and camera estimation network (1), the critic used in the adversarial training (2), and the reprojection part (3). 
The critic and the complete adversarial model are trained alternatingly as described above.

\subsection{Pose and Camera Estimation}
The pose and camera estimation network splits into two branches, one for regression of the pose and the other for the camera estimation.
In the following $\bm{X} \in \mathbb{R}^{3\times n}$ denotes a 3D human pose where each column contains the $xyz$-coordinates of a body joint.
In the neural network this matrix is written as a $3n$ dimensional vector.
Correspondingly, if $n$ joints are reconstructed the input of the pose and camera estimation network is a $2n$ dimensional vector containing the coordinates of the detected joints in the image.

The pose estimation part consists of two consecutive residual blocks, where each block has two hidden layers of 1000 densely connected neurons. 
For the activation functions we use leaky ReLUs \cite{leakyrelu2015} which produced the best results in our experiments.
The last layer outputs a $3n$ dimensional vector which contains the 3D pose and can be reshaped to $\bm{X}$.
The camera estimation branch has a similar structure as the pose estimation branch with the output being a $6$ dimensional vector containing the camera parameters.
Here, we use a weak perspective camera model that can be defined by only six variables.
To obtain the camera matrix the output vector is reshaped to $\bm{K} \in \mathbb{R}^{2\times 3}$.

\subsection{Reprojection Layer}
The reprojecting layer takes the output pose $\bm{X}$ of the 3D generator network and the camera $\bm{K}$ of the camera estimation network.
The reprojecting into 2D coordinate space can then be performed by
\begin{equation}
\label{eqn:rep}
	\bm{W}'=\bm{K}\bm{X}
	,
\end{equation}
where $\bm{W}'$ is called the \textit{2D reprojection} in the following.
This allows for the definition of a reprojection loss function
\begin{equation}
\mathcal{L}_{rep}(\bm{X},\bm{K})=
\| \bm{W} - \bm{K}\bm{X} \|_F
,
\end{equation}
where $\bm{W}$ is the input 2D pose observation matrix which has the same structure as $\bm{W}'$.
$\| \cdot \|_F$ denotes the Frobenius norm.
Note that the reprojection layer is a single layer which only performs the reprojection and does not have any trainable parameters.
To deal with occlusions columns in $\bm{W}$ and $\bm{X}$ that correspond to not detected joints can be set to zero.
This means they will have no influence on the value of the loss function.
The missing joints will then be hallucinated by the pose generator network according to the critic network.
In fact, the stacked hourglass network that produces the 2D joint detections \cite{StackedHourglassNewell2016} that we use as the input does not predict the spine joint.
We therefore set the corresponding columns in $\bm{W}$ and $\bm{X}$ to zero in all our experiments.

\subsection{Critic Network}
\begin{figure}[t!]
	\centering
	\includegraphics[width=0.45\textwidth]{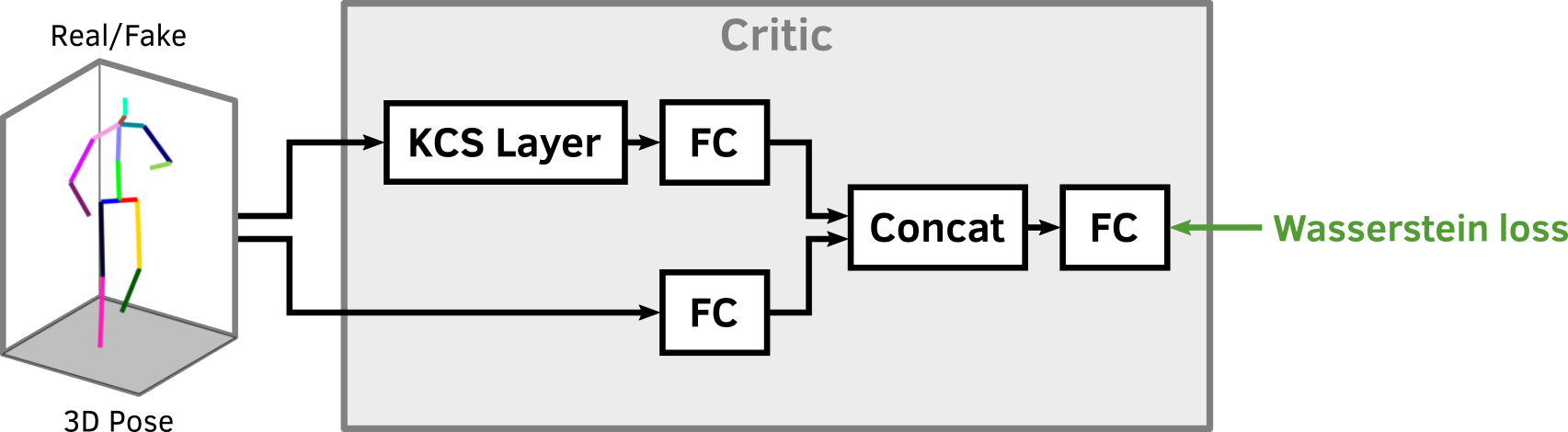}
	\caption{Network structure of the critic network. In the upper path the 3D pose is transformed into the KCS matrix and fed into a fully connected (FC) network. The lower path is build from multiple FC layers. The feature vectors of both paths are concatenated and fed into another FC layer which outputs the critic value.}
	\label{fig:critic}
\end{figure}
The complete network in Fig.~\ref{fig:network} is trained alternatingly with the critic network. 
The loss on the last layer of the critic is a Wasserstein loss function \cite{wgan2017}. 
The obvious choice of a critic network is a fully connected network with a structure similar to the pose regression network.
However, such networks struggle to detect properties of human poses such as kinematic chains, symmetry and joint angle limits.
Therefore, we follow the idea of Wandt et al. \cite{WanAck2018a} and add their \textit{kinematic chain space (KCS)} into our model.
We develop a KCS layer with a successive fully connected network which is added in parallel to the fully connected path.
These two paths in the critic network are merged before the output layer.
Fig.~\ref{fig:critic} shows the network structure of the critic.

The KCS matrix is a representation of a human pose containing joint angles and bone lenghts and can be computed by only two matrix multiplications.
A bone $\bm{b}_k$ is defined as the vector between the $r$-th and $t$-th joint
\begin{equation}
	\label{eqn_bone_diff}
	\bm{b}_k=\bm{p}_r-\bm{p}_t=\bm{X}\bm{c}
	,
\end{equation}
where
\begin{equation}
	\bm{c}=(0,\dots, 0, 1, 0, \dots, 0 ,-1 ,0 ,\dots,0)^T
	,
\end{equation}
with $1$ at position $r$ and $-1$ at position $t$. 
Note that the length of the vector $\bm{b}_k$ has the same direction and length as the corresponding bone. 
By concatenating $b$ bones a matrix $\bm{B} \in \mathbb{R}^{3\times b}$ can be defined as
\begin{equation}
	\label{eqn_bone}
	\bm{B}=(\bm{b}_1, \bm{b}_2, \dots, \bm{b}_b)
	.
\end{equation}
This leads to a matrix $\bm{C} \in \mathbb{R}^{j\times b}$ 
The matrix $\bm{B}$ is calculated by concatenating the corresponding vectors $\bm{c}$.
It follows
\begin{equation}
	\label{eqn_kin_space}
	\bm{B}=\bm{X}\bm{C}
	.
\end{equation}
Multiplying $\bm{B}$ with its transpose we compute the so called \textit{KCS~matrix}
\begin{equation}
	\label{eqn_Psi}
	\bm{\Psi}=\bm{B}^T\bm{B}=
	\begin{pmatrix}
		l_1^2 & \cdot & \cdot & \cdot \\
		\cdot & l_2^2 & \cdot & \cdot \\
		\cdot & \cdot & \ddots & \cdot \\
		\cdot & \cdot & \cdot & l_b^2 \\
	\end{pmatrix}
	.
\end{equation}
Because each entry in $\bm{\Psi}$ is an inner product of two bone vectors the KCS matrix has the bone lengths on its diagonal and a (scaled) angular representation on the other entries.
In contrast to an euclidean distance matrix \cite{Moreno_cvpr2017} the KCS matrix $\bm{\Psi}$ is easily calculated by two matrix multiplications.
This allows for an efficient implementation as an additional layer.
By giving the discriminator network an additional feature matrix it does not need to learn joint lengths computation and angular constraints on its own.
In fact, in our experiments it was not possible to achieve an acceptable symmetry between the left and right side of the body without the KCS matrix.
Section~\ref{sec:h36meval} shows how the 3D reconstruction benefits from adding the additional KCS layer.
In our experiments there was no difference between adding convolutional layers or fully connected layers after the KCS layer.
In the following we will use two fully connected layers, each containing 100 neurons, after the KCS layer.
Combined with the parallel fully connected network this leads to the critic structure in Fig.~\ref{fig:critic}.

\subsection{Camera}
Since the camera estimation sub-network in Fig.~\ref{fig:network} can produce any 6-dimensional vector we need to force the network to produce matrices describing weak perspective cameras. 
If the 3D poses and the 2D poses are centered at their root joint the camera matrix $\bm{K}$ projects $\bm{X}$ to $\bm{W}'$ according to Eq.~\ref{eqn:rep}. 
A weak perspective projection matrix $\bm{K}$ has the property 
\begin{equation}
\bm{K}\bm{K}^T=s^2\bm{I}_2
,
\end{equation}
where $s$ is the scale of the projection and $\bm{I}_2$ is the $2\times 2$ indentity matrix. 
Since the scale $s$ is unknown we derive a computationally efficient method of calculating $s$.
The scale $s$ equals to the largest singular value (or the $\ell_2$-norm) of $\bm{K}$.
Both singular values are equal.
Since the trace of $\bm{K}\bm{K}^T$ is the sum of the squared singular values
\begin{equation}
    s = \sqrt{trace(\bm{K}\bm{K}^T)/2}.
\end{equation}
The loss function can now be defined as 
\begin{equation}
\mathcal{L}_{cam}=
\| \frac{2}{trace(\bm{K}\bm{K}^T)}\bm{K}\bm{K}^T - \bm{I}_2 \|_F
,
\end{equation}
where $\|\cdot\|_F$ denotes the Frobenius norm.
Note that only one matrix multiplication is necessary to compute the quadratic scale.

\subsection{Data Preprocessing}
The camera estimation network infers the parameters of the weak perspective camera.
That means the camera matrix contains a rotational and a scaling component.
To avoid ambiguities between the camera and 3D pose rotation all the rotational and scaling components from the 3D poses are removed.
This is done by aligning every 3D pose to a template pose.
We do this by calculating the ideal rotation and scale for the corresponding shoulder and hip joints via procrustes alignment.
The resulting transformation is applied to all joints.

Depending on the persons size in the image the 2D joint detections can have arbitrary scale.
To remove the scale component we divide each 2D pose vector by its standard deviation.
Note that using this scaling technique the same person can have different sized 2D pose representations depending on the camera and 3D pose.
However, the value for all possible 2D poses is constrained.
The remaining scale variations are compensated by the cameras scale component.
In contrast to \eg \cite{martinez_2017_3dbaseline} we do not need to know the mean and standard deviation of the training set.
This allows for an easy transfer of our method to a different domain of 2D poses.

\subsection{Training}
We implemented the Improved Wasserstein GAN training procedure of \cite{iwgan2017}. 
In our experience this results in better and faster convergence compared to the traditional Wasserstein GAN \cite{wgan2017} and standard GAN training \cite{Goodfellow2014} using binary cross entropy or similar loss functions.
We use an initial learning rate of $0.001$ with exponential decay every $10$ epochs.

\section{Experiments}
\begin{figure*}[htp]
	\centering
	\includegraphics[width=0.9\textwidth]{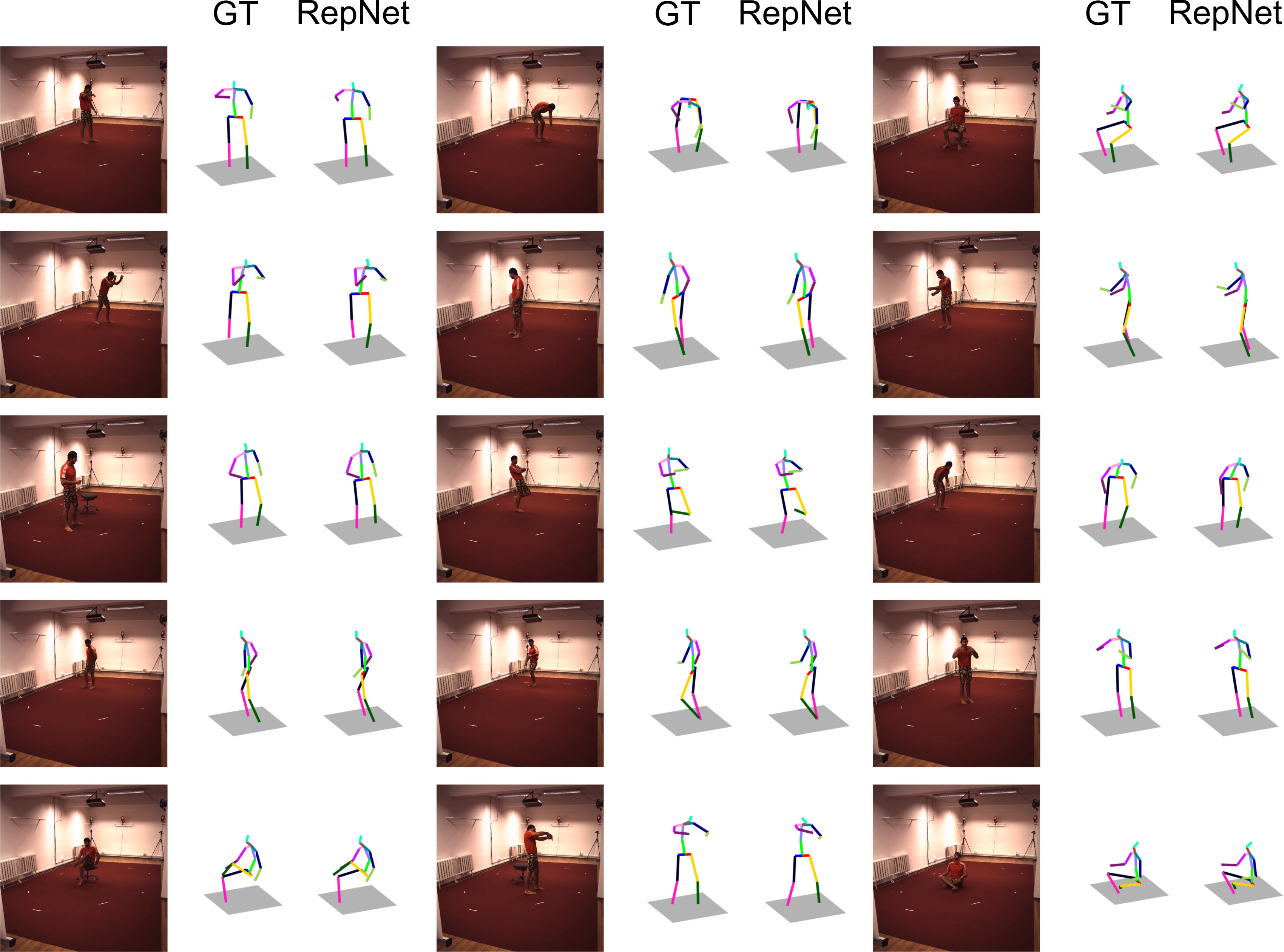}
	\caption{One example reconstruction for every motion from the test set of Human3.6M. The left 3D skeleton is the ground truth (GT) and the right shows our reconstruction (RepNet). Even difficult poses such as crossed legs or sitting on the floor are reconstructed well.}
	\label{fig:h36m_rec_results}
\end{figure*}
We perform experiments on the three datasets Human3.6M \cite{h36m_pami}, MPI-INF-3DHP \cite{mpii3dhp2017}, and LSP \cite{LeadsSports2010}.
Human3.6M is the largest benchmark dataset containing images temporally aligned to 2D and 3D correspondences.
Unless otherwise noted we use the training set of Human3.6M for training our networks.
To show quantitative results on unseen data we evaluate our method on MPI-INF-3DHP.
For unusual poses and camera angles subjective results are shown on LSP.
Matching most comparable methods we use stacked hourglass networks \cite{StackedHourglassNewell2016} for 2D joint estimations from the input images in most of the experiments.

\subsection{Quantitative Evaluation on Human3.6M}
\label{sec:h36meval}
\begin{table*}[htp]
	\footnotesize
	\caption{Results for the reconstruction of the Human3.6M dataset compared to other state-of-the-art methods following \textit{Protocol-I} (no ridig alignment). All numbers are taken from the referenced papers. For comparison the row \textit{RepNet+2DGT} shows the error when using the ground truth 2D labels. The column \textit{WS} denotes weakly supervised approaches. Note that there are no results available for other weakly supervised works.}
	\centering
	\resizebox{0.98\textwidth}{!}{
	\begin{tabular}{l|c|ccccccccccccccc|c}
 		Protocol-I & WS & Direct. & Disc. & Eat & Greet & Phone & Photo & Pose & Purch. & Sit & SitD & Smoke & Wait & Walk & WalkD & WalkT & Avg. \\
		\hline
		\rowcolor{grayLight}
		LinKDE  \cite{Ionescu14} & & 132.7 & 183.6 & 132.3 & 164.4 & 162.1 & 205.9 & 150.6 & 171.3 & 151.6 & 243.0 & 162.1 & 170.7 & 177.1 & 96.6 & 127.9 & 162.1 \\
		\rowcolor{grayDark}
Tekin et al . \cite{Tekin2016} & & 102.4 & 147.2 & 88.8 & 125.3 & 118.0 & 182.7 & 112.4 & 129.2 & 138.9 & 224.9 & 118.4 & 138.8 & 126.3 & 55.1 & 65.8 & 125.0\\
\rowcolor{grayLight}
Zhou et al . \cite{Zhou2016} & & 87.4 & 109.3 & 87.1 & 103.2 & 116.2 & 143.3 & 106.9 & 99.8 & 124.5 & 199.2 & 107.4 & 118.1 & 114.2 & 79.4 & 97.7 & 113.0\\
\rowcolor{grayDark}
Du et al. \cite{Du2016} & & 85.1 & 112.7 & 104.9 & 122.1 & 139.1 & 135.9 & 105.9 & 166.2 & 117.5 & 226.9 & 120.0 & 117.7 & 137.4 & 99.3 & 106.5 & 126.5\\
\rowcolor{grayLight}
Park et al. \cite{Park2016} & & 100.3 & 116.2 & 90.0 & 116.5 & 115.3 & 149.5 & 117.6 & 106.9 & 137.2 & 190.8 & 105.8 & 125.1 & 131.9 & 62.6 & 96.2 & 117.3\\
\rowcolor{grayDark}
Zhou et al. \cite{zhou2016deep} & & 91.8 & 102.4 & 96.7 & 98.8 & 113.4 & 125.2 & 90.0 & 93.8 & 132.2 & 159.0 & 107.0 & 94.4 & 126.0 & 79.0 & 99.0 & 107.3\\
\rowcolor{grayLight}
Luo et al. \cite{OriNet2018} & & 68.4 & 77.3 & 70.2 & 71.4 & 75.1 & 86.5 & 69.0 & 76.7 & 88.2 & 103.4 & 73.8 & 72.1 & 83.9 & 58.1 & 65.4 & 76.0 \\
		\rowcolor{grayDark}
		    Pavlakos et al. \cite{Pavlakos2017} & & 67.4 & 71.9 & 66.7 & 69.1 & 72.0 & 77.0 & 65.0 & 68.3 & 83.7 & 96.5 & 71.7 & 65.8 & 74.9 & 59.1 & 63.2 & 71.9\\
		    \rowcolor{grayLight}
		    Zhou et al. \cite{Zhou_2017_ICCV} & & 54.8 & 60.7 & 58.2 & 71.4 & 62.0 & 65.5 & 53.8 & 55.6 & 75.2 & 111.6 & 64.2 & 66.1 & 63.2 & 51.4 & 55.3 & 64.9 \\
		\rowcolor{grayDark}
		    Martinez et al. \cite{martinez_2017_3dbaseline} & & 53.3 & 60.8 & 62.9 & 62.7 & 86.4 & 82.4 & 57.8 & 58.7 & 81.9 & 99.8 & 69.1 & 63.9 & 50.9 & 67.1 & 54.8 & 67.5 \\
		\hline
		\hline
		RepNet (Ours) & \checkmark & 77.5 & 85.2 & 82.7 & 93.8 & 93.9 & 101.0 & 82.9 & 102.6 & 100.5 & 125.8 & 88.0 & 84.8 & 72.6 & 78.8 & 79.0 & 89.9 \\
		RepNet+2DGT (Ours) & \checkmark & 50.0 & 53.5 & 44.7 & 51.6 & 49.0 & 58.7 & 48.8 & 51.3 & 51.1 & 66.0 & 46.6 & 50.6 & 42.5 & 38.8 & 60.4& 50.9 \\
	\end{tabular}}
	\label{tab:protocol1results}
\end{table*} 

\begin{table*}[htp]
	\caption{Results for the reconstruction of the Human3.6M dataset compared to other state-of-the-art methods following \textit{Protocol-II} (rigid alignment). All numbers are taken from the referenced papers, except rows marked with * that are taken from \cite{AIGN2017}. Although we do not improve over supervised methods on this specific dataset our method clearly outperforms all other weakly supervised approaches (column \textit{WS}). The best results for the weakly supervised methods are marked in bold. The second best approach that is not ours is underlined. For comparison the last row \textit{RepNet+2DGT} shows the error when using the ground truth 2D labels.}
	\centering
	\resizebox{0.98\textwidth}{!}{
	\begin{tabular}{l|c|ccccccccccccccc|c}
 		Protocol-II & WS & Direct. & Disc. & Eat & Greet & Phone & Photo & Pose & Purch. & Sit & SitD & Smoke & Wait & Walk & WalkD & WalkT & Avg. \\
		\hline
		\rowcolor{grayLight}
		Akther and Black \cite{Akhter2015} &  & 199.2 & 177.6 & 161.8 & 197.8 & 176.2 & 186.5 & 195.4 & 167.3 & 160.7 & 173.7 & 177.8 & 181.9 & 198.6 & 176.2 & 192.7 & 181.1\\
		\rowcolor{grayDark}
		Ramakrishna et al. \cite{Ramakrishna12} &  & 37.4 & 149.3 & 141.6 & 154.3 & 157.7 & 158.9 & 141.8 & 158.1 & 168.6 & 175.6 & 160.4 & 161.7 & 174.8  & 150.0 & 150.2 & 157.3 \\
		\rowcolor{grayLight}
		Zhou et al. \cite{ZhouConvexRelax2016} &  & 99.7 & 95.8 & 87.9 & 116.8 & 108.3 & 107.3 & 93.5 & 95.3 & 109.1 & 137.5 & 106.0 & 102.2 & 110.4 & 106.5 & 115.2 & 106.7 \\
		\rowcolor{grayDark}
		Bogo et al. \cite{Bogo:ECCV:2016} &  &  62.0 & 60.2 & 67.8 & 76.5 & 92.1 & 77.0 & 73.0 & 75.3 & 100.3 & 137.3 & 83.4 & 77.3 & 79.7 & 86.8 & 87.7 & 82.3\\
		\rowcolor{grayLight}
		Moreno-Noguer \cite{Moreno_cvpr2017} &  & 66.1 & 61.7 & 84.5 & 73.7 & 65.2 & 67.2 & 60.9 & 67.3 & 103.5 & 74.6 & 92.6 & 69.6 & 78.0 & 71.5 & 73.2 & 74.0 \\
		\rowcolor{grayDark}
		Martinez et al. \cite{martinez_2017_3dbaseline} & & 44.8 & 52.0 & 44.4 & 50.5 & 61.7 & 59.4 & 45.1 & 41.9 & 66.3 & 77.6 & 54.0 & 58.8 & 35.9 & 49.0 & 40.7 & 52.1 \\
		\rowcolor{grayLight}
		Luo et al. \cite{OriNet2018} & & 40.8 & 44.6 & 42.1 & 45.1 & 48.3 & 54.6 & 41.2 & 42.9 & 55.5 & 69.9 & 46.7 & 42.5 & 36.0 & 48.0 & 41.4 & 46.6	\\
		\hline
		\hline
		\rowcolor{grayLight}
		3Dinterpreter* \cite{3dinterpreter2016} & \checkmark & 78.6 & \underline{90.8} & 92.5 & 89.4 & 108.9 & 112.4 & 77.1 & \underline{106.7} & 127.4 & 139.0 & 103.4 & 91.4 & 79.1 & - & - & 98.4 \\
		\rowcolor{grayDark}
		AIGN \cite{AIGN2017} & \checkmark & \underline{77.6} & 91.4 & \underline{89.9} & \underline{88.0} & \underline{107.3} & \underline{110.1} & \underline{75.9} & 107.5 & \underline{124.2} & \underline{137.8} & \underline{102.2} & \underline{90.3} & \underline{78.6} & - & - & \underline{97.2} \\
		\hline
		RepNet (Ours) & \checkmark & \bfseries 53.0 &\bfseries 58.3 &\bfseries 59.6 &\bfseries 66.5 &\bfseries 72.8 &\bfseries 71.0 &\bfseries 56.7 & \bfseries 69.6 & \bfseries 78.3 & \bfseries 95.2 & \bfseries 66.6 & \bfseries 58.5 & \bfseries 63.2 & \bfseries 57.5 & \bfseries 49.9 & \bfseries 65.1 \\
		RepNet-noKCS (Ours) & \checkmark & 63.1 & 67.4 & 71.5 & 78.5 & 85.9 & 82.6 & 70.8 & 82.7 & 92.2 & 116.6 & 77.6 & 72.2 & 65.3 & 73.2 & 69.6 & 77.9 \\
		RepNet+2DGT (Ours) & \checkmark & 33.6 & 38.8 & 32.6 & 37.5 & 36.0 & 44.1 & 37.8 & 34.9 & 39.2 & 52.0 & 37.5 & 39.8 & 34.1 & 40.3 & 34.9 & 38.2  \\
	\end{tabular}}
	\label{tab:protocol2results}
\end{table*} 
In the literature there are two main evaluation protocols on the Human3.6M dataset using subjects 1, 5, 6, 7, 8 for training and subject 9, 11 for testing.
Both protocols calculate the \textit{mean per joint positioning error} (MPJPE), i.e. the mean euclidean distance between the reconstructed and the ground truth joint coordinates.
Protocol-I computes the MPJPE directly whereas protocol-II first employs a rigid alignment between the poses.
For a sequence the MPJPE's are summed and divided by the number of frames.

Table~\ref{tab:protocol1results} shows the results of protocol-I without a rigid alignment.
The rotation of the pose relative to the camera can be directly calculated from the camera matrix estimated by the camera regression network.
Rotating the reconstructed pose in the world frame of the dataset gives the final 3D pose.
Table~\ref{tab:protocol2results} shows the results of protocol-II using a rigid alignment before calculating the error.
The row \textit{RepNet-noKCS} shows the errors without using the KCS layer.
It can be seen that the additional KCS layer in the discriminator significantly improves the pose estimation.
We are aware of the fact that our method will not be able to outperform supervised methods trained to perform exceptionally well on Human3.6M, such as \cite{martinez_2017_3dbaseline} and \cite{OriNet2018}.
Instead, in this section we show that even if we ignore the 2D-3D correspondences and train weakly supervised our network achieves comparable results to supervised state-of-the-art methods and is even better than most of them.
Comparing to weakly supervised approaches \cite{3dinterpreter2016,AIGN2017} we outperform the best by about 30\% on protocol-II.
For subjective evaluation the 1500th frame for every motion can be seen in Fig.~\ref{fig:h36m_rec_results}.
For comparability we show the same frame from every motion sequence from the same viewing angle.
Even difficult poses, for instance sitting cross-legged, are reconstructed well.

In our opinion, although widely used on Human3.6M, the euclidean distance is not the only metric that should be considered to evaluate the performance of a human pose estimation system.
Since there are some single frames that cannot be reconstructed well and can be seen as outliers we also calculate the median of the MPJPE over all frames.
Additionally, we calculate the \textit{percentage of correctly positioned keypoints} (PCK3D) as defined by \cite{mpii3dhp2017} in Table~\ref{tab:pck_h36m}.

\begin{table}[htp]
	\footnotesize
	\caption{Performance of our method regarding the median and PCK3D errors for the Human3.6M dataset.}
	\centering
	\begin{tabular}{l|cccc}
 		 & mean & median & PCK3D \\
		\hline
		\rowcolor{grayLight}
		RepNet  & 65.1 & 60.0 & 93.0 \\
		\rowcolor{grayDark}
		RepNet+2DGT  & 38.2 & 36.0 & 98.6 \\
	\end{tabular}
	\label{tab:pck_h36m}
\end{table}

In the following section we will show that although we do not improve on all supervised state-of-the-art methods directly trained on Human3.6M our approach outperforms every other known method on MPI-INF-3DHP without additional training.

\subsection{Quantitative Evaluation on MPI-INF-3DHP}
\label{sec:3dhp}
\begin{table}[htp]
	\footnotesize
	\caption{Results for the MPI-INF-3DHP dataset. All numbers are taken from the referenced papers, except the row marked with * which is taken from \cite{singleshotmultiperson2018}. Without training on this dataset the proposed method outperforms every other method. The row \textit{RepNet 3DHP} shows the result when using the training set of MPI-INF-3DHP. The column \textit{WS} denotes weakly supervised approaches. A higher value is better for 3DPCK and AUC while a lower value is better for MPJPE. The best results are marked in bold and the second best approach is underlined.}
	\centering
	\begin{tabular}{l|c|ccc}
 		Method & WS & 3DPCK & AUC & MPJPE \\
		\hline
		\rowcolor{grayLight}
		Mehta et al. \cite{mpii3dhp2017} & & 76.5 & 40.8 & 117.6 \\
		\rowcolor{grayDark}
		VNect \cite{VNect_SIGGRAPH2017} &  & 76.6 & 40.4 & 124.7 \\
		\rowcolor{grayLight}
		LCR-Net\cite{lcrnet2017}* & & 59.6 & 27.6 & 158.4 \\		
		\rowcolor{grayDark}
		Zhou et al. \cite{Zhou_2017_ICCV} & & 69.2 & 32.5 & 137.1 \\
		\rowcolor{grayLight}
		Multi Person \cite{singleshotmultiperson2018} & & 75.2 & 37.8 & 122.2 \\
		\rowcolor{grayDark}
		OriNet \cite{OriNet2018} & & \underline{81.8} & 45.2 & \textbf{89.4} \\
		\rowcolor{grayLight}
		Kanazawa \cite{Kanazawa2018} & \checkmark & 77.1 &	40.7 & 113.2 \\
		\rowcolor{grayDark}
		Yang et al. \cite{Yang2018} & \checkmark & 69.0 & 32.0 & - \\
		\hline
		RepNet H36M (Ours)  & \checkmark & \underline{81.8}  & \underline{54.8} & \underline{92.5}\\
		RepNet 3DHP (Ours)  & \checkmark &   \textbf{82.5}   & \textbf{58.5}  & 97.8 \\
	\end{tabular}
	\label{tab:mpiinf3dhp}
\end{table} 

Our main contribution is a neural network that infers even unseen human poses while maintaining a meaningful 3D pose.
We compare our method against several state-of-the-art approaches.
Table~\ref{tab:mpiinf3dhp} shows the results for different metrics.
We clearly outperform every other method without having trained our model on this specific dataset.
Even approaches trained on the training set of MPI-INF-3DHP perform worse than ours.
This shows the generalization capability of our network.
The row \textit{RepNet 3DHP} is the result when training on the training set of MPI-INF-3DHP.
There is only a minor improvement of the 3DPCK and AUC and even a minor deterioration of the MPJPE compared to the network trained on Human3.6M.
This suggests that the critic network converges to a similar distribution of feasible human poses for both training sets.

\subsection{Plausibility of the Reconstructions}
\begin{table}[htp]
	\footnotesize
	\caption{Symmetry error in $mm$ of the reconstructed 3D poses on the different datasets with and without the KCS. Adding the KCS layer to the critic networks results in significantly more plausible poses.}
	\centering
	\begin{tabular}{l|cccc}
 		Method & mean & std & max \\
		\hline
		\rowcolor{grayLight}
		H36M noKCS  & 31.9 & 9.3  & 61.3 \\
		\rowcolor{grayDark}
		H36M KCS    & 8.2  & 3.8  & 20.5 \\
		\rowcolor{grayLight}
		3DHP noKCS  & 32.9  & 21.9  & 143.9   \\
		\rowcolor{grayDark}
		3DHP KCS    & 11.2  & 8.0  & 54.7 \\
	\end{tabular}
	\label{tab:symmerr}
\end{table}
The metrics used for evaluation in Sec.~\ref{sec:h36meval} and \ref{sec:3dhp} compare the estimated 3D pose to the ground truth.
However, a low error in this metrics is not necessarily an indication for a plausible human pose since the reconstructed pose can still violate joint angle limits or symmetries of the human body.
For this purpose we introduce a new metric based on bone length symmetry.
We calculate bone lengths of the lower and upper arms and legs since there is the largest error per joint.
By summing the absolute differences of all matching bones on the right and left side of the body we can calculate a \textit{symmetry error}.
The mean symmetry error of the ground truth poses from the test set of Human3.6M and MPI-INF-3DHP for all subjects is $0.7mm\pm 0.8mm$ (max. $2.6mm$) and $2.1mm\pm 1.3mm$ (max. $7.6mm$), respectively.
This leads us to the conclusion that an equality between the left and right side and therefore a low symmetry error is one reasonable metric for the plausibility of a human pose.
Table~\ref{tab:symmerr} compares several implementations of our network in terms of the symmetry error.
It can be clearly seen that the KCS layer has a significant impact on this metric.
The higher values for the MPI-INF-3DHP dataset can be explained by the larger differences in symmetry of the ground truth data.

\subsection{Noisy observations}
\begin{table*}[h!tp]
	\caption{Evaluation results for protocol-II (rigid aligment) with different levels of gaussian noise $\mathcal{N}(0,\sigma)$ ($\sigma$ is the standard deviation) added to the ground truth 2D positions (\textit{GT}). The 2D detector noise has large impact on the 3D reconstruction. The right three columns show the mean, standard deviation, and maximal symmetry error in millimeter.}
	\centering
	\resizebox{0.99\textwidth}{!}{
	\begin{tabular}{l|ccccccccccccccc|c|ccc}
	    \multicolumn{17}{c}{ } & \multicolumn{3}{c}{symmetry}\\
 		Protocol-II & Direct. & Disc. & Eat & Greet & Phone & Photo & Pose & Purch. & Sit & SitD & Smoke & Wait & Walk & WalkD & WalkT & Avg. & mean & std & max\\
		\hline
		\rowcolor{grayLight}
		GT & 33.6 & 38.8 & 32.6 & 37.5 & 36.0 & 44.1 & 37.8 & 34.9 & 39.2 & 52.0 & 37.5 & 39.8 & 34.1 & 40.3 & 34.9 & 38.2 & 6.2 & 3.7 & 20.8  \\
		\rowcolor{grayDark}
		GT + $\mathcal{N}(0,5)$ & 54.0 & 56.8 & 52.7 & 56.5 & 54.4 & 59.7 & 55.7 & 54.1 & 56.3 & 68.5 & 56.1 & 58.7 & 57.6 & 56.7 & 55.3 & 56.9  & 9.6 & 4.0 & 25.0\\
		\rowcolor{grayLight}
        GT + $\mathcal{N}(0,10)$ & 70.4 & 72.2 & 72.8 & 75.1 & 70.2 & 84.1 & 68.4 & 89.3 & 74.0 & 94.1 & 68.3 & 74.3 & 67.7 & 73.5 & 70.0 & 74.9  & 13.0 & 3.8 & 24.2  \\
        \rowcolor{grayDark}
        GT + $\mathcal{N}(0,15)$ & 86.3 & 88.0 & 87.5 & 89.9 & 84.0 & 98.1 & 84.0 & 104.2 & 87.4 & 107.7 & 82.3 & 89.3 & 85.1 & 89.0 & 86.0 & 89.9  & 17.6  & 4.2 & 32.1 \\
        \rowcolor{grayLight}
        GT + $\mathcal{N}(0,20)$ & 101.6 & 103.0 & 101.6 & 104.5 & 97.5 & 112.2 & 99.3 & 118.1 & 100.9 & 121.5 & 95.9 & 104.0 & 101.6 & 104.7 & 102.3 & 104.6  & 22.7 & 4.5  & 37.5 \\
	\end{tabular}}
	\label{tab:noise}
\end{table*} 
Since the performance of our network appears to depend a lot on the detections of the 2D pose detector we evaluate our network on different levels of noise.
Following \cite{Moreno_cvpr2017} we add gaussian noise $\mathcal{N}(0,\sigma)$ to the ground truth 2D joint positions, where $\sigma$ is the standard deviation in pixel. 
The results for Human3.6M under protocol-II are shown in Table~\ref{tab:noise}.
The error scales linearly with the standard deviation.
This indicates that the noise of the 2D joint detector has a major impact on the results.
Considering Tables~\ref{tab:protocol1results} and \ref{tab:protocol2results} an improved detector will enhance the results to a level where they outperform current state-of-the-art supervised approaches.

Please note that the maximum person size from head to toe is approximately $200$px in the input data.
Therefore, gaussian noise with a standard deviation of $\sigma=20$px can be considered as extremely large.
However, due to the critic network using the KCS layer the output of the pose estimation network is still a plausible human pose.
To demonstrate this we additionally investigated the average, standard deviation and maximal symmetry error for the different noise levels which is also shown in Table~\ref{tab:noise}.
As expected the error increases only slightly since the critic network enforces plausible human poses.
Even for noise levels as high as $\mathcal{N}(0,20)$ we achieve an average symmetry error of only $22.7mm \pm 4.5mm$ which can be considered as very low.


\subsection{Qualitative Evaluation}
\begin{figure*}[h!tp]
	\centering
	\includegraphics[width=0.9\textwidth]{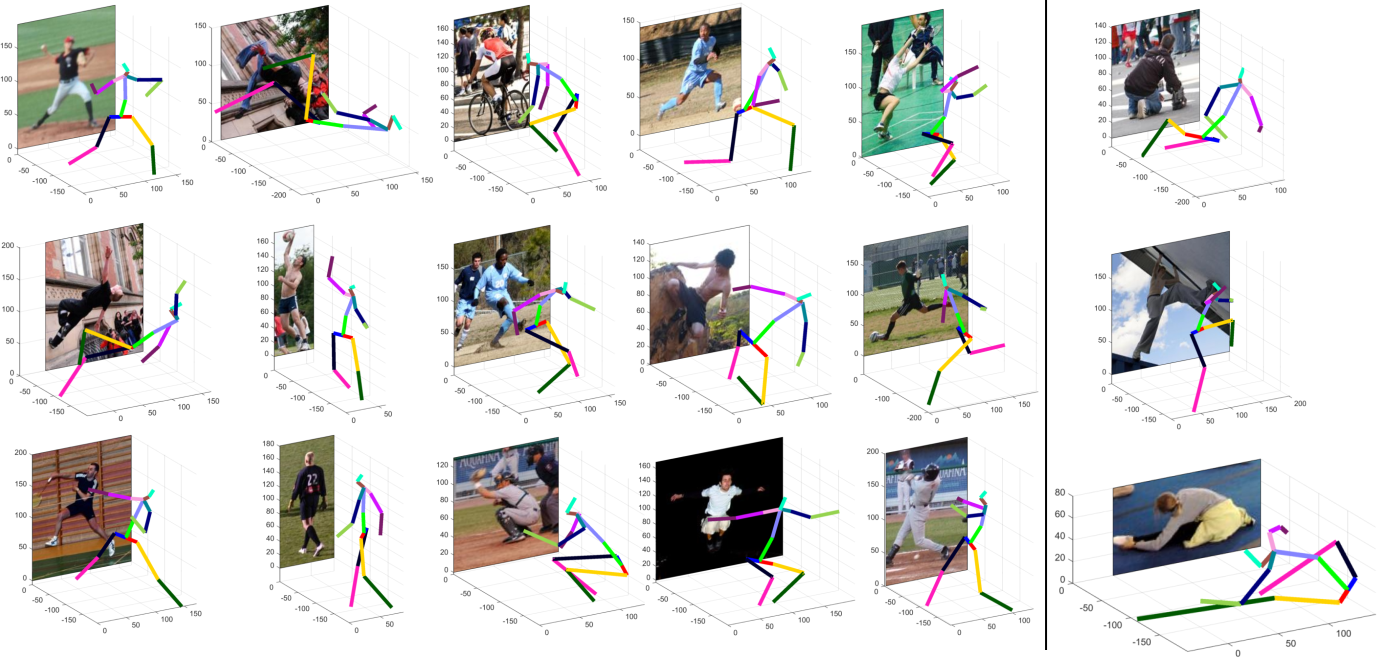}
	\caption{Example 3D pose estimations from the LSP dataset. Good reconstructions are in the left columns. The right column shows some failure cases with very unusual poses or camera angles. Although not perfect the poses are still plausible and close to the correct poses.}
	\label{fig:lsp_results}
	\vspace{-5px}
\end{figure*}
For a subjective evaluation we use the Leeds Sports Pose dataset (LSP) \cite{LeadsSports2010}.
This dataset contains 2000 images of different people doing sports.
There is a large variety in poses including stretched poses close to the limits of possible joint angles.
Some of these poses and camera angles were never seen before by our network.
Nevertheless, it is able to predict plausible 3D poses for most of the images.
Fig.~\ref{fig:lsp_results} shows some of the reconstructions achieved by our method.
There are many subjectively well reconstructed poses, even if these are extremely stretched and captured from uncommon camera angles.
Note that our network was only trained on the camera angles of Human3.6M.
This underlines that an understanding of plausible poses and 2D projections is learned.
The right column in Fig.~\ref{fig:lsp_results} shows some failure cases and emphasizes a limitation of this approach: poses or camera angles that are too different from the training data cannot be reconstructed well.
However, the reconstructions are still plausible human poses and in most cases at least near to the correct pose. 

\subsection{Computational Time}
We see our method as a building block in a larger image-to-3D points system.
Current state-of-the-art 2D keypoint detectors such as \cite{cao2017realtime} achieve real-time performance (approximately $100$ms per frame) on standard hardware.
Our network adds another $0.05$ms per frame and therefore has nearly no impact on the runtime.
Assuming the 2D keypoint detection takes no time we achieve a frame rate of $20000$fps on an Nvidia TITAN X.

\section{Conclusion}
This paper presented RepNet: a weakly supervised training method for a 3D human pose estimation neural network that infers 3D poses from 2D joint detections in single images.
We proposed to use an additional camera estimation network and our novel reprojection layer that projects the estimated 3D pose back to 2D.
By exploiting state-of-the-art techniques in neural network research, such as improved Wasserstein GANs \cite{iwgan2017} and kinematic chain spaces \cite{WanAck2018a}, we were able to develop a weakly supervised training procedure that does not need 2D to 3D correspondences.
This not only outperforms previous weakly supervised methods but also avoids overfitting of the network to a limited amount of training data.
We achieved state-of-the-art performance on the benchmark dataset Human3.6M, even compared to most supervised approaches.
Using the network trained on Human3.6M to predict 3D poses from the unseen data of the MPI-INF-3DHP dataset showed an improvement over all other methods.
We also performed a subjective evaluation on the LSP dataset where we achieved good reconstructions even on images with uncommon poses and perspectives.

{\small
\bibliographystyle{ieee}
\bibliography{literature}

\begin{thebibliography}{10}\itemsep=-1pt

\bibitem{Akhter2015}
I.~Akhter and M.~J. Black.
\newblock Pose-conditioned joint angle limits for {3D} human pose
  reconstruction.
\newblock In {\em IEEE Conf.~on Computer Vision and Pattern Recognition (CVPR
  2015)}, pages 1446--1455, June 2015.

\bibitem{AllKas2017}
T.~Alldieck, M.~Kassubeck, B.~Wandt, B.~Rosenhahn, and M.~Magnor.
\newblock Optical flow-based 3d human motion estimation from monocular video.
\newblock In {\em German Conference on Pattern Recognition (GCPR)}, Sept. 2017.

\bibitem{wgan2017}
M.~Arjovsky, S.~Chintala, and L.~Bottou.
\newblock Wasserstein generative adversarial networks.
\newblock In D.~Precup and Y.~W. Teh, editors, {\em Proceedings of the 34th
  International Conference on Machine Learning}, volume~70 of {\em Proceedings
  of Machine Learning Research}, pages 214--223, International Convention
  Centre, Sydney, Australia, 06--11 Aug 2017. PMLR.

\bibitem{Bogo:ECCV:2016}
F.~Bogo, A.~Kanazawa, C.~Lassner, P.~Gehler, J.~Romero, and M.~J. Black.
\newblock Keep it {SMPL}: Automatic estimation of {3D} human pose and shape
  from a single image.
\newblock In {\em Computer Vision -- ECCV 2016}, Lecture Notes in Computer
  Science. Springer International Publishing, Oct. 2016.

\bibitem{cao2017realtime}
Z.~Cao, T.~Simon, S.-E. Wei, and Y.~Sheikh.
\newblock Realtime multi-person 2d pose estimation using part affinity fields.
\newblock In {\em CVPR}, 2017.

\bibitem{ChenR17}
C.~Chen and D.~Ramanan.
\newblock 3d human pose estimation = 2d pose estimation + matching.
\newblock In {\em 2017 {IEEE} Conference on Computer Vision and Pattern
  Recognition, {CVPR} 2017}, pages 5759--5767, 2017.

\bibitem{ChenC09}
Y.-L. Chen and J.~Chai.
\newblock 3d reconstruction of human motion and skeleton from uncalibrated
  monocular video.
\newblock In H.~Zha, R.~I. Taniguchi, and S.~J. Maybank, editors, {\em Asian
  Conference on Computer Vision (ACCV)}, volume 5994 of {\em Lecture Notes in
  Computer Science}, pages 71--82. Springer, 2009.

\bibitem{Du2016}
Y.~Du, Y.~Wong, Y.~Liu, F.~Han, Y.~Gui, Z.~Wang, M.~Kankanhalli, and W.~Geng.
\newblock {Marker-less 3D human motion capture with monocular image sequence
  and height-maps}.
\newblock In {\em European Conference on Computer Vision}, pages 20--36.
  Springer, 2016.

\bibitem{Goodfellow2014}
I.~J. Goodfellow, J.~Pouget-Abadie, M.~Mirza, B.~Xu, D.~Warde-Farley, S.~Ozair,
  A.~Courville, and Y.~Bengio.
\newblock Generative adversarial nets.
\newblock In {\em Proceedings of the 27th International Conference on Neural
  Information Processing Systems - Volume 2}, NIPS'14, pages 2672--2680,
  Cambridge, MA, USA, 2014. MIT Press.

\bibitem{iwgan2017}
I.~Gulrajani, F.~Ahmed, M.~Arjovsky, V.~Dumoulin, and A.~C. Courville.
\newblock Improved training of wasserstein gans.
\newblock In I.~Guyon, U.~V. Luxburg, S.~Bengio, H.~Wallach, R.~Fergus,
  S.~Vishwanathan, and R.~Garnett, editors, {\em Advances in Neural Information
  Processing Systems 30}, pages 5767--5777. Curran Associates, Inc., 2017.

\bibitem{Gupta2014}
A.~Gupta, J.~Martinez, J.~J. Little, and R.~J. Woodham.
\newblock 3d pose from motion for cross-view action recognition via non-linear
  circulant temporal encoding.
\newblock {\em 2014 IEEE Conference on Computer Vision and Pattern
  Recognition}, pages 2601--2608, 2014.

\bibitem{leakyrelu2015}
K.~He, X.~Zhang, S.~Ren, and J.~Sun.
\newblock Delving deep into rectifiers: Surpassing human-level performance on
  imagenet classification.
\newblock In {\em Proceedings of the 2015 IEEE International Conference on
  Computer Vision (ICCV)}, ICCV '15, pages 1026--1034, Washington, DC, USA,
  2015. IEEE Computer Society.

\bibitem{h36m_pami}
C.~Ionescu, D.~Papava, V.~Olaru, and C.~Sminchisescu.
\newblock Human3.6m: Large scale datasets and predictive methods for 3d human
  sensing in natural environments.
\newblock {\em IEEE Transactions on Pattern Analysis and Machine Intelligence},
  36(7):1325--1339, jul 2014.

\bibitem{Ionescu14}
C.~Ionescu, D.~Papava, V.~Olaru, and C.~Sminchisescu.
\newblock Human3.6m: Large scale datasets and predictive methods for 3d human
  sensing in natural environments.
\newblock {\em IEEE Transactions on Pattern Analysis and Machine Intelligence},
  36(7):1325--1339, 2014.

\bibitem{Jiang2010}
H.~Jiang.
\newblock 3d human pose reconstruction using millions of exemplars.
\newblock {\em 2010 20th International Conference on Pattern Recognition},
  pages 1674--1677, 2010.

\bibitem{LeadsSports2010}
S.~Johnson and M.~Everingham.
\newblock Clustered pose and nonlinear appearance models for human pose
  estimation.
\newblock In {\em Proceedings of the British Machine Vision Conference}, 2010.
\newblock doi:10.5244/C.24.12.

\bibitem{Kanazawa2018}
A.~Kanazawa, M.~J. Black, D.~W. Jacobs, and J.~Malik.
\newblock End-to-end recovery of human shape and pose.
\newblock In {\em IEEE Conference on Computer Vision and Pattern Recognition
  (CVPR)}. IEEE Computer Society, 2018.

\bibitem{Lee1985}
H.-J. Lee and Z.~Chen.
\newblock Determination of 3d human body postures from a single view.
\newblock {\em Computer Vision, Graphics, and Image Processing}, 30(2):148 --
  168, 1985.

\bibitem{OriNet2018}
C.~Luo, X.~Chu, and A.~L. Yuille.
\newblock Orinet: {A} fully convolutional network for 3d human pose estimation.
\newblock In {\em British Machine Vision Conference 2018, {BMVC} 2018,
  Northumbria University, Newcastle, UK, September 3-6, 2018}, page~92, 2018.

\bibitem{martinez_2017_3dbaseline}
J.~Martinez, R.~Hossain, J.~Romero, and J.~J. Little.
\newblock A simple yet effective baseline for 3d human pose estimation.
\newblock In {\em ICCV}, 2017.

\bibitem{mpii3dhp2017}
D.~Mehta, H.~Rhodin, D.~Casas, P.~Fua, O.~Sotnychenko, W.~Xu, and C.~Theobalt.
\newblock Monocular 3d human pose estimation in the wild using improved cnn
  supervision.
\newblock In {\em 3D Vision (3DV), 2017 Fifth International Conference on}.
  IEEE, 2017.

\bibitem{singleshotmultiperson2018}
D.~Mehta, O.~Sotnychenko, F.~Mueller, W.~Xu, S.~Sridhar, G.~Pons-Moll, and
  C.~Theobalt.
\newblock Single-shot multi-person 3d pose estimation from monocular rgb.
\newblock In {\em 3D Vision (3DV), 2018 Sixth International Conference on}.
  IEEE, sep 2018.

\bibitem{VNect_SIGGRAPH2017}
D.~Mehta, S.~Sridhar, O.~Sotnychenko, H.~Rhodin, M.~Shafiei, H.-P. Seidel,
  W.~Xu, D.~Casas, and C.~Theobalt.
\newblock Vnect: Real-time 3d human pose estimation with a single rgb camera.
\newblock volume~36, 7 2017.

\bibitem{Moreno_cvpr2017}
F.~Moreno-Noguer.
\newblock 3d human pose estimation from a single image via distance matrix
  regression.
\newblock In {\em Proceedings of the Conference on Computer Vision and Pattern
  Recognition (CVPR)}, 2017.

\bibitem{StackedHourglassNewell2016}
A.~Newell, K.~Yang, and J.~Deng.
\newblock Stacked hourglass networks for human pose estimation.
\newblock In {\em {ECCV} {(8)}}, volume 9912 of {\em Lecture Notes in Computer
  Science}, pages 483--499. Springer, 2016.

\bibitem{NBF2018}
M.~Omran, C.~Lassner, G.~Pons-Moll, P.~V. Gehler, and B.~Schiele.
\newblock Neural body fitting: Unifying deep learning and model-based human
  pose and shape estimation.
\newblock In {\em 3DV}, Sept. 2018.

\bibitem{Park2016}
S.~Park, J.~Hwang, and N.~Kwak.
\newblock 3d human pose estimation using convolutional neural networks with 2d
  pose information.
\newblock In {\em Computer Vision - {ECCV} 2016 Workshops - Amsterdam, The
  Netherlands, October 8-10 and 15-16, 2016, Proceedings, Part {III}}, pages
  156--169, 2016.

\bibitem{Pavlakos2017}
G.~Pavlakos, X.~Zhou, K.~G. Derpanis, and K.~Daniilidis.
\newblock Coarse-to-fine volumetric prediction for single-image 3d human pose.
\newblock {\em 2017 IEEE Conference on Computer Vision and Pattern Recognition
  (CVPR)}, pages 1263--1272, 2017.

\bibitem{pavlakos2018}
G.~Pavlakos, L.~Zhu, X.~Zhou, and K.~Daniilidis.
\newblock Learning to estimate 3{D} human pose and shape from a single color
  image.
\newblock In {\em CVPR}, 2018.

\bibitem{Ramakrishna12}
V.~Ramakrishna, T.~Kanade, and Y.~A. Sheikh.
\newblock Reconstructing 3d human pose from 2d image landmarks.
\newblock In {\em European Conference on Computer Vision (ECCV)}, October 2012.

\bibitem{Hossain2018}
M.~Rayat Imtiaz~Hossain and J.~J. Little.
\newblock Exploiting temporal information for 3d human pose estimation.
\newblock In {\em The European Conference on Computer Vision (ECCV)}, September
  2018.

\bibitem{lcrnet2017}
G.~Rogez, P.~Weinzaepfel, and C.~Schmid.
\newblock {LCR-Net: Localization-Classification-Regression for Human Pose}.
\newblock In {\em {CVPR 2017 - IEEE Conference on Computer Vision \& Pattern
  Recognition}}, pages 1216--1224, Honolulu, United States, July 2017. {IEEE}.

\bibitem{SimoSerraRATM12}
E.~Simo-Serra, A.~Ramisa, G.~Alenyà, C.~Torras, and F.~Moreno-Noguer.
\newblock Single image 3d human pose estimation from noisy observations.
\newblock In {\em Conference on Computer Vision and Pattern Recognition
  (CVPR)}, pages 2673--2680. IEEE, 2012.

\bibitem{Tekin2016}
B.~Tekin, A.~Rozantsev, V.~Lepetit, and P.~Fua.
\newblock Direct prediction of 3d body poses from motion compensated sequences.
\newblock In {\em 2016 {IEEE} Conference on Computer Vision and Pattern
  Recognition, {CVPR} 2016, Las Vegas, NV, USA, June 27-30, 2016}, pages
  991--1000, 2016.

\bibitem{AIGN2017}
H.~F. Tung, A.~W. Harley, W.~Seto, and K.~Fragkiadaki.
\newblock Adversarial inverse graphics networks: Learning 2d-to-3d lifting and
  image-to-image translation from unpaired supervision.
\newblock In {\em 2017 IEEE International Conference on Computer Vision
  (ICCV)}, pages 4364--4372, Oct 2017.

\bibitem{Tung2017}
H.-Y. Tung, H.-W. Tung, E.~Yumer, and K.~Fragkiadaki.
\newblock Self-supervised learning of motion capture.
\newblock In I.~Guyon, U.~V. Luxburg, S.~Bengio, H.~Wallach, R.~Fergus,
  S.~Vishwanathan, and R.~Garnett, editors, {\em Advances in Neural Information
  Processing Systems 30}, pages 5236--5246. Curran Associates, Inc., 2017.

\bibitem{Marcard2016}
T.~v.~Marcard, G.~Pons-Moll, and B.~Rosenhahn.
\newblock Human pose estimation from video and imus.
\newblock {\em IEEE Transactions on Pattern Analysis and Machine Intelligence},
  38(8):1533--1547, Aug 2016.

\bibitem{Marcard2018}
T.~von Marcard, R.~Henschel, M.~J. Black, B.~Rosenhahn, and G.~Pons-Moll.
\newblock Recovering accurate {3D} human pose in the wild using {IMUs} and a
  moving camera.
\newblock In {\em European Conference on Computer Vision (ECCV)}, volume
  Lecture Notes in Computer Science, vol 11214, pages 614--631. Springer, Cham,
  Sept. 2018.

\bibitem{Marcard2017}
T.~von Marcard, B.~Rosenhahn, M.~J. Black, and G.~Pons-Moll.
\newblock Sparse inertial poser: Automatic 3d human pose estimation from sparse
  imus.
\newblock {\em Computer Graphics Forum}, 36(2):349--360, 2017.

\bibitem{Wandt2016}
B.~Wandt, H.~Ackermann, and B.~Rosenhahn.
\newblock 3d reconstruction of human motion from monocular image sequences.
\newblock {\em IEEE Transactions on Pattern Analysis and Machine Intelligence},
  38(8):1505--1516, 2016.

\bibitem{WanAck2018a}
B.~Wandt, H.~Ackermann, and B.~Rosenhahn.
\newblock A kinematic chain space for monocular motion capture.
\newblock In {\em ECCV Workshops}, Sept. 2018.

\bibitem{Wang2014}
C.~Wang, Y.~Wang, Z.~Lin, A.~Yuille, and W.~Gao.
\newblock Robust estimation of 3d human poses from a single image.
\newblock In {\em Conference on Computer Vision and Pattern Recognition
  (CVPR)}, 2014.

\bibitem{Wei2009}
X.~K. Wei and J.~Chai.
\newblock Modeling 3d human poses from uncalibrated monocular images.
\newblock In {\em 2009 IEEE 12th International Conference on Computer Vision},
  pages 1873--1880, Sept 2009.

\bibitem{3dinterpreter2016}
J.~Wu, T.~Xue, J.~J. Lim, Y.~Tian, J.~B. Tenenbaum, A.~Torralba, and W.~T.
  Freeman.
\newblock Single image 3d interpreter network.
\newblock In {\em European Conference on Computer Vision (ECCV)}, 2016.

\bibitem{Yang2018}
W.~Yang, W.~Ouyang, X.~Wang, J.~Ren, H.~Li, and X.~Wang.
\newblock 3d human pose estimation in the wild by adversarial learning.
\newblock In {\em CVPR}, 2018.

\bibitem{ZelWan2017}
P.~Zell, B.~Wandt, and B.~Rosenhahn.
\newblock Joint 3d human motion capture and physical analysis from monocular
  videos.
\newblock In {\em The IEEE Conference on Computer Vision and Pattern
  Recognition (CVPR) Workshops}, July 2017.

\bibitem{Zhou_2017_ICCV}
X.~Zhou, Q.~Huang, X.~Sun, X.~Xue, and Y.~Wei.
\newblock Towards 3d human pose estimation in the wild: A weakly-supervised
  approach.
\newblock In {\em The IEEE International Conference on Computer Vision (ICCV)},
  Oct 2017.

\bibitem{zhou2016deep}
X.~Zhou, X.~Sun, W.~Zhang, S.~Liang, and Y.~Wei.
\newblock Deep kinematic pose regression.
\newblock pages 186--201, 2016.

\bibitem{ZhouConvexRelax2016}
X.~Zhou, M.~Zhu, S.~Leonardos, and K.~Daniilidis.
\newblock Sparse representation for 3d shape estimation: A convex relaxation
  approach.
\newblock {\em IEEE Transactions on Pattern Analysis and Machine Intelligence},
  39(8):1648--1661, Aug 2017.

\bibitem{Zhou2016}
X.~Zhou, M.~Zhu, S.~Leonardos, K.~G. Derpanis, and K.~Daniilidis.
\newblock Sparseness meets deepness: 3d human pose estimation from monocular
  video.
\newblock In {\em The IEEE Conference on Computer Vision and Pattern
  Recognition (CVPR)}, June 2016.

\end{thebibliography}
}

\end{document}